\documentclass[11pt,a4paper]{article}
\usepackage[hyperref]{conll2018}
\usepackage{times}
\usepackage{latexsym}
\usepackage{url}
\usepackage{graphicx}
\usepackage{amsmath}
\usepackage{mathtools}
\usepackage{multirow}
\graphicspath{{Figures/}}
\usepackage{tabularx,tabulary}
\usepackage{subcaption}
\usepackage{url}
\usepackage{amssymb}
\usepackage[linesnumbered,ruled]{algorithm2e}
\usepackage{diagbox}
\usepackage{xcolor}
\usepackage{pifont}
\usepackage{adjustbox}
\aclfinalcopy 

\newcommand{\misplaced}[1]{\textcolor{violet}{#1}}
\newcommand{\missing}[1]{\textcolor{red}{#1}}
\newcommand{\redundant}[1]{\textcolor{orange}{#1}}
\newcommand{\wrong}[1]{\textcolor{teal}{#1}}
\newcommand{\spelling}[1]{\textcolor{olive}{#1}}
\newcommand{\OK}[1]{\textcolor{blue}{#1}}
\newcommand*{\affaddr}[1]{#1}
\newcommand*{\affmark}[1][*]{\textsuperscript{#1}}
\newcommand*{\email}[1]{\texttt{#1}}

\title{Dual Latent Variable Model for Low-Resource Natural Language Generation in Dialogue Systems}

\author{
Van-Khanh Tran\affmark[1,2] and Le-Minh Nguyen\affmark[1] \\
\affaddr{\affmark[1]Japan Advanced Institute of Science and Technology, JAIST\\
					1-1 Asahidai, Nomi, Ishikawa, 923-1292, Japan}\\
\email{\{tvkhanh, nguyenml\}@jaist.ac.jp}\\
\affaddr{\affmark[2]University of Information and Communication Technology, ICTU\\
	Thai Nguyen University, Vietnam}\\
\email{tvkhanh@ictu.edu.vn}
}

\begin{document}
\maketitle
\begin{abstract}
Recent deep learning models have shown improving results to natural language generation (NLG) irrespective of providing sufficient annotated data. 
However, a modest training data may harm such models' performance. 
Thus, how to build a generator that can utilize as much of knowledge from a low-resource setting data is a crucial issue in NLG. 
This paper presents a variational neural-based generation model to tackle the NLG problem of having limited labeled dataset, in which we integrate a variational inference into an encoder-decoder generator and introduce a novel auxiliary auto-encoding with an effective training procedure.
Experiments showed that the proposed methods not only outperform the previous models when having sufficient training dataset but also show strong ability to work acceptably well when the training data is scarce.
\end{abstract}

\section{Introduction}
Natural language generation (NLG) plays an critical role in Spoken dialogue systems (SDSs) with the NLG task is mainly to convert a meaning representation produced by the dialogue manager, \textit{i.e.}, dialogue act (DA), into natural language responses. SDSs are typically developed for various specific domains, \textit{i.e.}, flight reservations \cite{levin2000t}, buying a tv or a laptop \cite{wensclstm15}, searching for a hotel or a restaurant \cite{wenthwsjy15}, and so forth. Such systems  often require well-defined ontology datasets that are extremely time-consuming and expensive to collect. There is, thus, a need to build NLG systems that can work acceptably well when the training data is in short supply.

There are two potential solutions for above-mentioned problems, which are \textit{domain adaptation} training and \textit{model designing for low-resource} training. First, \textit{domain adaptation} training which aims at learning from sufficient source domain a model that can perform acceptably well on a different target domain with a limited labeled target data.  Domain adaptation generally involves two different types of datasets, one from a source domain and the other from a target domain. Despite providing promising results for low-resource setting problems, the methods still need an adequate training data at the source domain site.

Second, \textit{model designing for low-resource setting} has not been well studied in the NLG literature. 
The generation models have achieved very good performances irrespective of providing sufficient labeled datasets \cite{wensclstm15,wenthwsjy15,tran-nguyen-tojo:2017:W17-55,tran-nguyen:2017:CoNLL}. 
However, small training data easily result in worse generation models in the supervised learning methods. Thus, this paper presents an explicit way to construct an effective low-resource setting generator.

In summary, we make the following contributions, in which we: 
(i) propose a variational approach for an NLG problem which benefits the generator to not only outperform the previous methods when there is a sufficient training data but also perform acceptably well regarding low-resource data;
(ii) present a variational generator that can also adapt faster to a new, unseen domain using a limited amount of in-domain data; 
(iii) investigate the effectiveness of the proposed method in different scenarios, including ablation studies, scratch, domain adaptation, and semi-supervised training with varied proportion of dataset.

\section{Related Work}
\label{sec:related_work}
Recently, the RNN-based generators have shown improving results in tackling the NLG problems in task oriented-dialogue systems with varied proposed methods, such as HLSTM \cite{wenthwsjy15}, SCLSTM \cite{wensclstm15}, or especially RNN Encoder-Decoder models integrating with attention mechanism, such as Enc-Dec \cite{wentoward}, and RALSTM \cite{tran-nguyen:2017:CoNLL}. However, such models have proved to work well only when providing a sufficient in-domain data since a modest dataset may harm the models' performance. 

In this context, one can think of a potential solution where the domain adaptation learning is utilized. The source domain, in this scenario, typically contains a sufficient amount of annotated data such that a model can be efficiently built, while there is often little or no labeled data in the target domain. A phrase-based statistical generator \cite{Mairesse:2010:PSL:1858681.1858838} using graphical models and active learning, and a multi-domain procedure \cite{wen2016multi} via data counterfeiting and discriminative training. 
However, a question still remains as how to build a generator that can directly work well on a scarce dataset.

Neural variational framework for generative models of text have been studied extensively. 
\newcite{chung2015recurrent} proposed a recurrent latent variable model for sequential data by integrating latent random variables into hidden state of an RNN. 
A hierarchical multi scale recurrent neural networks was proposed to learn both hierarchical and temporal representation \cite{chung2016hierarchical}, while \newcite{DBLP:journals/corr/BowmanVVDJB15} presented a variational autoencoder for unsupervised generative language model. 
\newcite{sohn2015learning} proposed a deep conditional generative model for structured output prediction, whereas \newcite{2016arXiv160507869Z} introduced a variational neural machine translation that incorporated a continuous latent variable to model underlying semantics of sentence pairs. 
To solve the exposure-bias problem \cite{bengio2015scheduled} \newcite{zhang2017deconvolutional,shen2017deconvolutional} proposed a seq2seq purely convolutional and deconvolutional autoencoder, \newcite{yang2017improved} proposed to use a dilated CNN decoder in a latent-variable model, or \newcite{semeniuta2017hybrid} proposed a hybrid VAE architecture with convolutional and deconvolutional components.

\section{Dual Latent Variable Model} \label{sec:duallvm}
\begin{figure}[!ht] 
	\centering
    \includegraphics[width=0.35\textwidth, height=2.7cm]{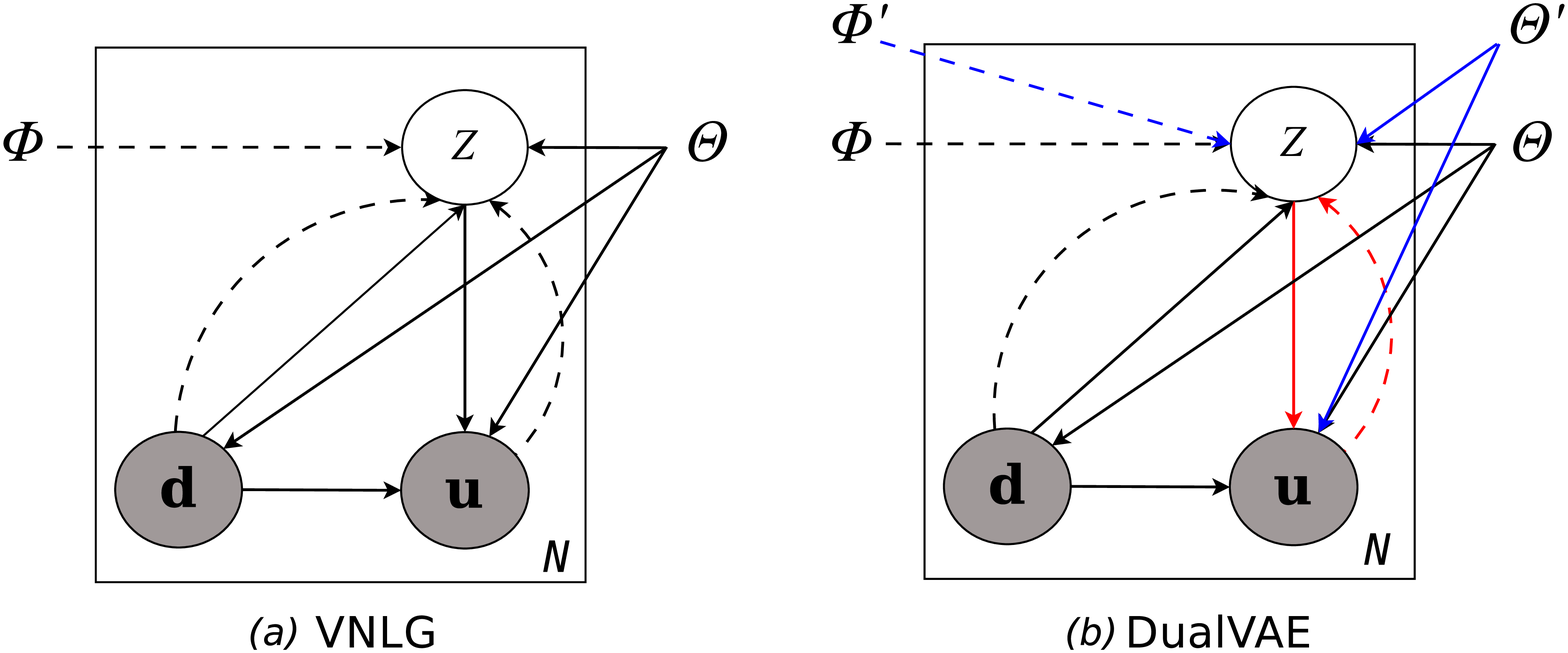} \vspace{-10pt}
    \caption{Illustration of proposed variational models as a directed graph. \textit{(a)} VNLG: joint learning both variational parameters $\phi$ and generative model parameters $\theta$. \textit{(b)} DualVAE: \textcolor{red}{red} and \textcolor{blue}{blue} arrows form a standard VAE (parameterized by $\phi'$ and $\theta'$) as an auxiliary auto-encoding to the VNLG model denoted by \textcolor{red}{red} and black arrows.}
    \label{fig:DualVAE-GM} \vspace{-15pt}
\end{figure}
\subsection{Variational Natural Language Generator}
We make an assumption about the existing of a continuous latent variable $z$ from a underlying semantic space of DA-Utterance pairs $(\textbf{d}, \textbf{u})$, so that we explicitly model the space together with variable $\textbf{d}$ to guide the generation process, \textit{i.e.}, $p(\textbf{u}|z, \textbf{d})$. 
The original conditional probability $p(\textbf{y}|\textbf{d})$ modeled by a vanilla encoder-decoder network is thus reformulated as follows:
\begin{equation}\label{eq:p-y-d-z}
p(\textbf{u}|\textbf{d})=\int_zp(\textbf{u},z|\textbf{d})\textbf{d}_z=\int_zp(\textbf{u}|z,\textbf{d})p(z|\textbf{d})\textbf{d}_z
\end{equation}
This latent variable enables us to model the underlying semantic space as a global signal for generation. However, the incorporating of latent variable into the probabilistic model arises two difficulties in \textit{(i)} modeling the intractable posterior inference $p(z|\textbf{d}, \textbf{u})$ and \textit{(ii)} whether or not the latent variables $z$ can be modeled effectively in case of low-resource setting data. 

To address the difficulties, we propose an encoder-decoder based variational model to natural language generation (VNLG) by integrating a variational autoencoder \cite{kingma2013auto} into an encoder-decoder generator \cite{tran-nguyen:2017:CoNLL}. Figure \ref{fig:DualVAE-GM}-\textit{(a)} shows a graphical model of VNLG.
We then employ deep neural networks to approximate the prior $p(z|\textbf{d})$, true posterior $p(z|\textbf{d}, \textbf{u})$, and decoder $p(\textbf{u}|z,\textbf{d})$. 
To tackle the first issue, the intractable posterior is approximated from both the DA and utterance information $q_\phi(z|\textbf{d}, \textbf{u})$ under the above assumption. 
In contrast, the prior is modeled to condition on the DA only $p_\theta(z|\textbf{d})$ due to the fact that the DA and utterance of a training pair usually share the same semantic information, \textit{i.e.}, a given DA \textit{inform}(name=`\textit{ABC}'; area=`\textit{XYZ}') contains key information of the corresponding utterance ``The hotel \textit{ABC} is in \textit{XYZ} area''. 
The underlying semantic space with having more information encoded from both the prior and the posterior provides the generator a potential solution to tackle the second issue.
Lastly, in generative process, given an observation DA \textbf{d} the output \textbf{u} is generated by the decoder network $p_\theta(\textbf{u}|z,\textbf{d})$ under the guidance of the global signal $z$ which is drawn from the prior distribution $p_\theta(z|\textbf{d})$. According to \cite{sohn2015learning}, the variational lower bound can be recomputed as:
\begin{equation}\label{eq:lowerbound}
\begin{aligned}
\mathcal{L}(\theta, &\phi, \textbf{d}, \textbf{u}) = -KL(q_\phi(z|\textbf{d}, \textbf{u}) || p_\theta(z|\textbf{d})) 	\\
+ &\mathbb{E}_{q_\phi(z|\textbf{d}, \textbf{u})}[\log p_\theta(\textbf{u}|z, \textbf{d})] \leq \log p(\textbf{u}|\textbf{d})
\end{aligned}
\end{equation}


\subsubsection{Variational Encoder Network} 
The encoder consists of two networks: (\textit{i}) a Bidirectional LSTM (BiLSTM) which encodes the sequence of slot-value pairs $\{{\textbf{sv}}_i\}^{T_{DA}}_{i=1}$ by separate parameterization of slots and values \cite{wentoward}; and (\textit{ii}) a shared CNN/RNN Utterance Encoder which encodes the corresponding utterance. 
The encoder network, thus, produces both the DA representation $\textbf{h}_\textbf{D}$ and the utterance representation $\textbf{h}_\textbf{U}$ vectors which flow into the inference and decoder networks, and the posterior approximator, respectively (see Suppl. 1.1).

\subsubsection{Variational Inference Network }
This section models both the prior $p_\theta(z|\textbf{d})$ and the posterior $q_\phi(z|\textbf{d},\textbf{u})$ by utilizing neural networks.

\textbf{Neural Posterior Approximator}: We approximate the intractable posterior distribution of $z$ to simplify the posterior inference, in which we first projects both DA and utterance representations onto the latent space:\vspace{-5pt}
\begin{equation}\label{eq:approximation-f}
\textbf{h}'_z = g(\textbf{W}_z[\textbf{h}_\textbf{D};\textbf{h}_\textbf{U}]+b_z)\vspace{-5pt}
\end{equation}
where $\textbf{W}_z\in \mathbb{R}^{d_z \times (d_{\textbf{h}_\textbf{D}}+d_{\textbf{h}_\textbf{U}})}$, $b_z \in \mathbb{R}^{d_z}$ are matrix and bias parameters respectively, $d_z$ is the dimensionality of the latent space, and we set $g(.)$ to be ReLU in our experiments. We then approximate the posterior as:\vspace{-5pt}
\begin{equation}\label{eq:approximation-form}
q_\phi(z|\textbf{d},\textbf{u}) = \mathcal{N}(z; \mu_1(\textbf{h}'_z), \sigma_1^2(\textbf{h}'_z)\textbf{\textit{I}})
\end{equation}
with mean $\mu_1$ and standard variance $\sigma_1$ are the outputs of the neural network as follows:
\begin{equation}\label{eq:posterior}
\mu_1 = \textbf{W}_{\mu_1}\textbf{h}'_z + b_{\mu_1}, \log \sigma_1^2=\textbf{W}_{\sigma_1}\textbf{h}'_z+b_{\sigma_1}
\end{equation}
where $\mu_1$, $\log\sigma_1^2$ are both $d_z$ dimension vectors.

\begin{figure}[!ht] 
	\centering
    \includegraphics[width=0.4\textwidth, height=.4\textwidth]{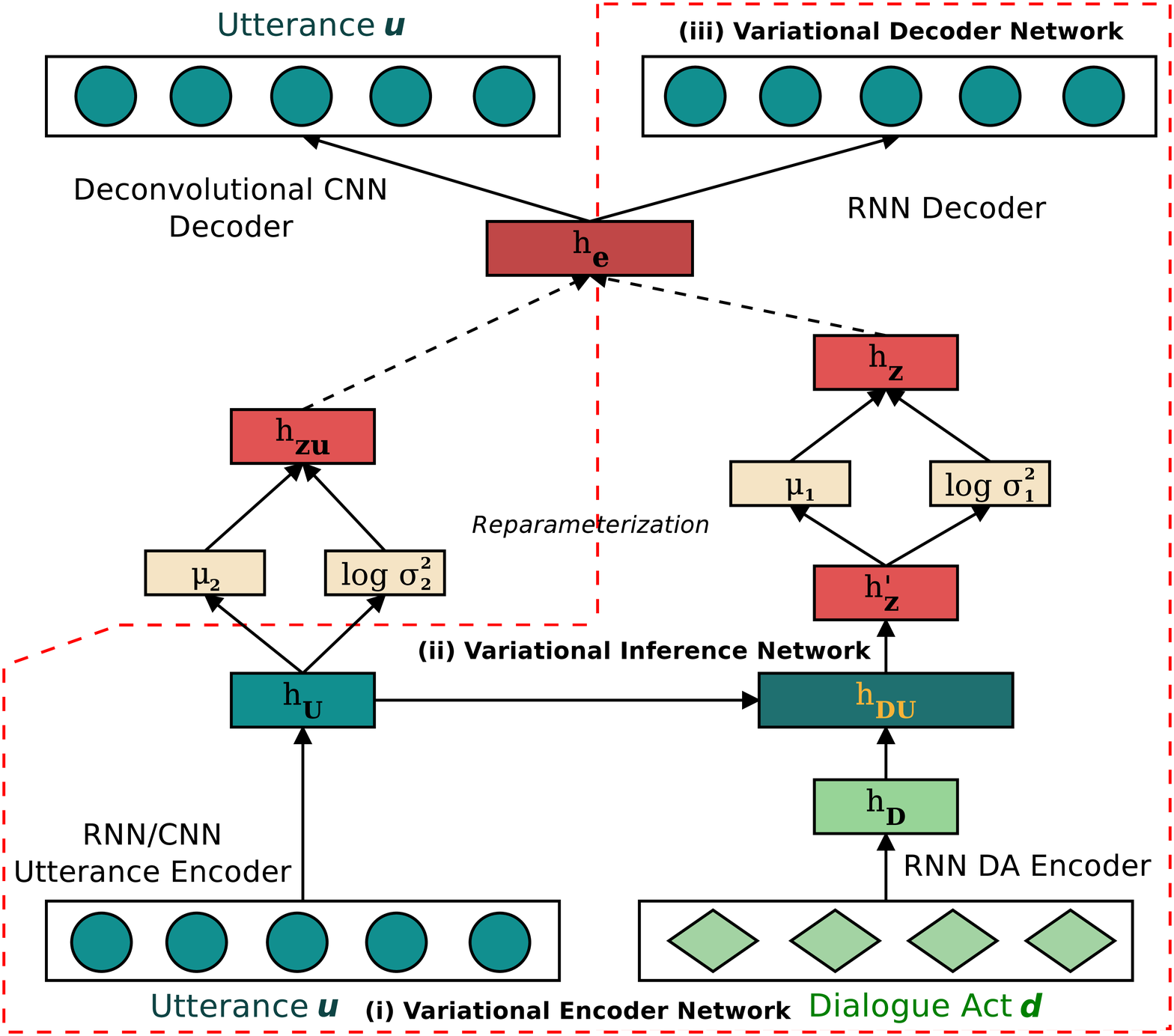} 
    \caption{The Dual latent variable model consists of two VAE models: \textit{(i)} a VNLG (red-dashed box) is to generate utterances and \textit{(ii)} a Variational CNN-DCNN is an auxiliary auto-encoding model (left side). The RNN/CNN Utterance Encoder is shared between the two VAEs.}
    \label{fig:dualvae}
     
\end{figure}

\textbf{Neural Prior}: We model the prior as follows:
\begin{equation}\label{eq:prior}
p_{\theta}(z|\textbf{d}) = \mathcal{N}(z; \mu'_1(\textbf{d}), {\sigma'}^2_1(\textbf{d})\textbf{\textit{I}})
\end{equation}
where $\mu_1'$ and $\sigma_1'$ of the prior are neural models only based on the Dialogue Act representation, which are the same as those of the posterior $q_\phi(z|\textbf{d},\textbf{u})$ in Eq. \ref{eq:approximation-f} and \ref{eq:posterior}, except for the absence of $\textbf{h}_\textbf{U}$. To obtain a representation of the latent variable $z$, we re-parameterize it as follows:
$\textbf{h}_z = \mu_1 + \sigma_1 \odot \epsilon$ where $\epsilon \sim \mathcal{N}(0, \textbf{\textit{I}}).$

Note here that the parameters for the prior and the posterior are independent of each other. Moreover, during decoding we set $\textbf{h}_z$ to be the mean of the prior $p_\theta(z|\textbf{d})$, \textit{i.e.}, $\mu_1'$ due to the absence of the utterance $\textbf{u}$. 
In order to integrate the latent variable $\textbf{h}_z$ into the decoder, we use a non-linear transformation to project it onto the output space for generation:
$\textbf{h}_e = g(\textbf{W}_e\textbf{h}_z + b_e) \refstepcounter{equation}(\theequation)\label{eq:he-1}$, where $\textbf{h}_e \in \mathbb{R}^{d_e}$.

\subsubsection{Variational Decoder Network}
Given a DA $\textbf{d}$ and the latent variable $z$, the decoder calculates the probability over the generation $\textbf{u}$ as a joint probability of ordered conditionals: 
\begin{equation}\label{eq:lstm-decoder}
p(\textbf{u}|z,\textbf{d})=\prod_{t=1}^{T_\textbf{U}} p(\textbf{u}_t|\textbf{u}_{<t}, z, \textbf{d})
\end{equation}
where $p(\textbf{u}_t|\textbf{u}_{<t}, z, \textbf{d})$=$g'(\textnormal{RALSTM}(\textbf{u}_{t}, \textbf{h}_{t-1}, \textbf{d}_{t})$. The RALSTM cell \cite{tran-nguyen:2017:CoNLL} is slightly modified in order to integrate the representation of latent variable, \textit{i.e.}, $\textbf{h}_e$, into the computational cell (see Suppl. 1.3), in which the latent variable can affect the hidden representation through the gates. This allows the model can indirectly take advantage of the underlying semantic information from the latent variable $z$. In addition, when the model learns unseen dialogue acts, the semantic representation $\textbf{h}_e$ can benefit the generation process (see Table \ref{tab:cVariational-DualVAE-results}).

We finally obtain the VNLG model with RNN Utterance Encoder (R-VNLG) or with CNN Utterance Encoder (C-VNLG).

\subsection{Variational CNN-DCNN Model}
This standard VAE model (left side in Figure \ref{fig:dualvae}) acts as an auxiliary auto-encoding for utterance (used at training time) to the VNLG generator. The model consists of two components. While the \textit{shared} CNN Utterance Encoder with the VNLG model is to compute the latent representation vector $\textbf{h}_\textbf{U}$ (see Suppl. 1.1.3), a Deconvolutional CNN Decoder to decode the latent representation $\textbf{h}_e$ back to the source text (see Suppl. 2.1). Specifically, after having the vector representation $\textbf{h}_\textbf{U}$, we apply another linear regression to obtain the distribution parameter $\mu_2 = \textbf{W}_{\mu_2}\textbf{h}_\textbf{U} + b_{\mu_2}$ and $\log \sigma_2^2=\textbf{W}_{\sigma_2}\textbf{h}_\textbf{U}+b_{\sigma_2}$.
We then re-parameterize them to obtain a latent representation $\textbf{h}_{zu} = \mu_2 + \sigma_2 \odot \epsilon$, where $\epsilon \sim \mathcal{N}(0, \textbf{\textit{I}})$.
In order to integrate the latent variable $\textbf{h}_{zu}$ into the DCNN Decoder, we use the \textit{shared} non-linear transformation as in Eq. \ref{eq:he-1} (denoted by the black-dashed line in Figure \ref{fig:dualvae}) as: $\textbf{h}_e = g(\textbf{W}_e\textbf{h}_{zu} + b_e)$.

The entire resulting model, named DualVAE, by incorporating the VNLG with the Variational CNN-DCNN model, is depicted in Figure \ref{fig:dualvae}.

\section{Training Dual Latent Variable Model}

\subsection{Training VNLG Model}
Inspired by work of \newcite{2016arXiv160507869Z}, we also employ the Monte-Carlo method to approximate the expectation of the posterior in Eq. \ref{eq:lowerbound}, \textit{i.e}. $\mathbb{E}_{q_{\phi}(z|\textbf{d}, \textbf{u})}[.] \simeq \frac{1}{M}\sum_{m=1}^{M} \log p_{\theta}(\textbf{u}|\textbf{d}, \textbf{h}_z^{(m)})$ where $M$ is the number of samples. In this work, the joint training objective $\mathcal{L}_{\textnormal{VNLG}}$ for a training instance pair $(\textbf{d}, \textbf{u})$ is formulated as:
\begin{equation}\label{eq:g-objective-function}
\begin{aligned}
&\mathcal{L}(\theta, \phi, \textbf{d}, \textbf{u}) \simeq -KL(q_{\phi}(z|\textbf{d},\textbf{u})||p_\theta(z|\textbf{d})) \\
&+ \frac{1}{M}\sum_{m=1}^{M} \sum_{t=1}^{T_{U}}\log p_{\theta}(\textbf{u}_t|\textbf{u}_{<t}, \textbf{d}, \textbf{h}_z^{(m)})
\end{aligned}
\end{equation}
where $\textbf{h}_z^{(m)} = \mu + \sigma \odot \epsilon^{(m)}$, and $\epsilon^{(m)} \sim \mathcal{N}(0, \textbf{I})$, and $\theta$ and $\phi$ denote decoder and encoder parameters, respectively. The first term is the KL divergence between two Gaussian distribution, and the second term is the approximation expectation. We simply set $M=1$ which degenerates the second term to the objective of conventional generator. Since the objective function in Eq. \ref{eq:g-objective-function} is differentiable, we can jointly optimize the parameter $\theta$ and variational parameter $\phi$ using standard gradient ascent techniques.
However, the KL divergence loss tends to be significantly small during training \cite{DBLP:journals/corr/BowmanVVDJB15}. As a results, the decoder does not take advantage of information from the latent variable $z$. Thus, we apply the KL cost annealing strategy that encourages the model to encode meaningful representations into the latent vector $z$, in which we gradually anneal the KL term from $0$ to $1$. This helps our model to achieve solutions with non-zero KL term.

\subsection{Training Variational CNN-DCNN Model}
The objective function $\mathcal{L}_{\textnormal{CNN-DCNN}}$ of the Variational CNN-DCNN model is the standard VAE lower bound and maximized as follows:
\begin{equation}
\begin{aligned}
\mathcal{L}(\theta'&, \phi', \textbf{u}) = -KL(q_{\phi'}(z|\textbf{u}) || p_{\theta'}(z)) \\ 
&+ \mathbb{E}_{q_{\phi'}(z|\textbf{u})}[\log p_{\theta'}(\textbf{u}|z)] \leq \log p(\textbf{u})
\end{aligned}                                    
\end{equation}
where $\theta'$ and $\phi'$ denote decoder and encoder parameters, respectively. 
During training, we also consider a denoising autoencoder where we slightly modify the input by swapping some arbitrary word pairs.

\subsection{Joint Training Dual VAE Model}
To allow the model explore and balance maximizing the variational lower bound between the Variational CNN-DCNN model and VNLG model, an objective is joint training as follows:
\begin{equation}
\mathcal{L}_\textnormal{\textbf{DualVAE}} = \mathcal{L}_{\textnormal{VNLG}} + \alpha\mathcal{L}_{\textnormal{CNN-DCNN}} 
\end{equation}
where $\alpha$ controls the relative weight between two variational losses. During training, we anneal the value of $\alpha$ from $1$ to $0$, so that the dual latent variable learned can gradually focus less on reconstruction objective of the CNN-DCNN model, only retain those features that are useful for the generation objective.

\subsection{Joint Cross Training Dual VAE Model}
To allow the dual VAE model explore and encode useful information of the Dialogue Act into the latent variable, we further take a cross training between two VAEs by simply replacing the RALSTM Decoder of the VNLG model with the DCNN Utterance Decoder and its objective training $\mathcal{L}_{\textnormal{DA-DCNN}}$ as:
\begin{equation}\label{eq:cross-objective-function}
\begin{aligned}
\mathcal{L}(\theta', \phi, \textbf{d}, \textbf{u}) &\simeq -KL(q_{\phi}(z|\textbf{d},\textbf{u})||p_{\theta'}(z|\textbf{d})) \\
&+ \mathbb{E}_{q_{\phi}(z|\textbf{d}, \textbf{u})}[\log p_{\theta'}(\textbf{u}|z, \textbf{d})],
\end{aligned}
\end{equation}
and a joint cross training objective is employed:
\begin{equation}
\begin{aligned}
\mathcal{L}_\textnormal{\textbf{CrossVAE}} &= \mathcal{L}_{\textnormal{VNLG}} \\ 
&+ \alpha(\mathcal{L}_{\textnormal{CNN-DCNN}} + \mathcal{L}_{\textnormal{DA-DCNN}})
\end{aligned}
\end{equation}

\section{Experiments}
We assessed the proposed models on four different original NLG domains: finding a restaurant and hotel \cite{wenthwsjy15}, or buying a laptop and television \cite{wentoward}.

\subsection{Evaluation Metrics and Baselines}\label{subsec:evaluation-metrics}
The generator performances were evaluated using the two metrics: the BLEU and the slot error rate ERR by adopting code from an NLG toolkit\footnotemark. We compared the proposed models against strong baselines which have been recently published as NLG benchmarks of those datasets, including (i) gating models such as HLSTM \cite{wenthwsjy15}, and SCLSTM \cite{wensclstm15}; and (ii) attention models such as Enc-Dec \cite{wentoward}, RALSTM \cite{tran-nguyen:2017:CoNLL}.
\footnotetext{https://github.com/shawnwun/RNNLG}

\subsection{Experimental Setups}
In this work, the CNN Utterance Encoder consists of $L=3$ layers, which for a sentence of length $T=73$, embedding size $d=100$, stride length $s=\{2,2,2\}$, number of filters $k=\{300, 600, 100\}$ with filter sizes $h=\{5,5,16\}$, results in feature maps \textbf{V} of sizes $\{35 \times 300, 16 \times 600, 1 \times 100\}$, in which the last feature map corresponds to latent representation vector $\textbf{h}_\textbf{U}$.

The hidden layer size and beam width were set to be $100$ and $10$, respectively, and the models were trained with a $70\%$ of keep dropout rate. We performed $5$ runs with different random initialization of the network, and the training process is terminated by using early stopping. For the variational inference, we set the latent variable size to be $300$. We used Adam optimizer with the learning rate is initially set to be $0.001$, and after $5$ epochs the learning rate is decayed every epoch using an exponential rate of $0.95$.

\begin{figure}[!ht] 
	\centering
    \includegraphics[width=0.38\textwidth, height=3.7cm]{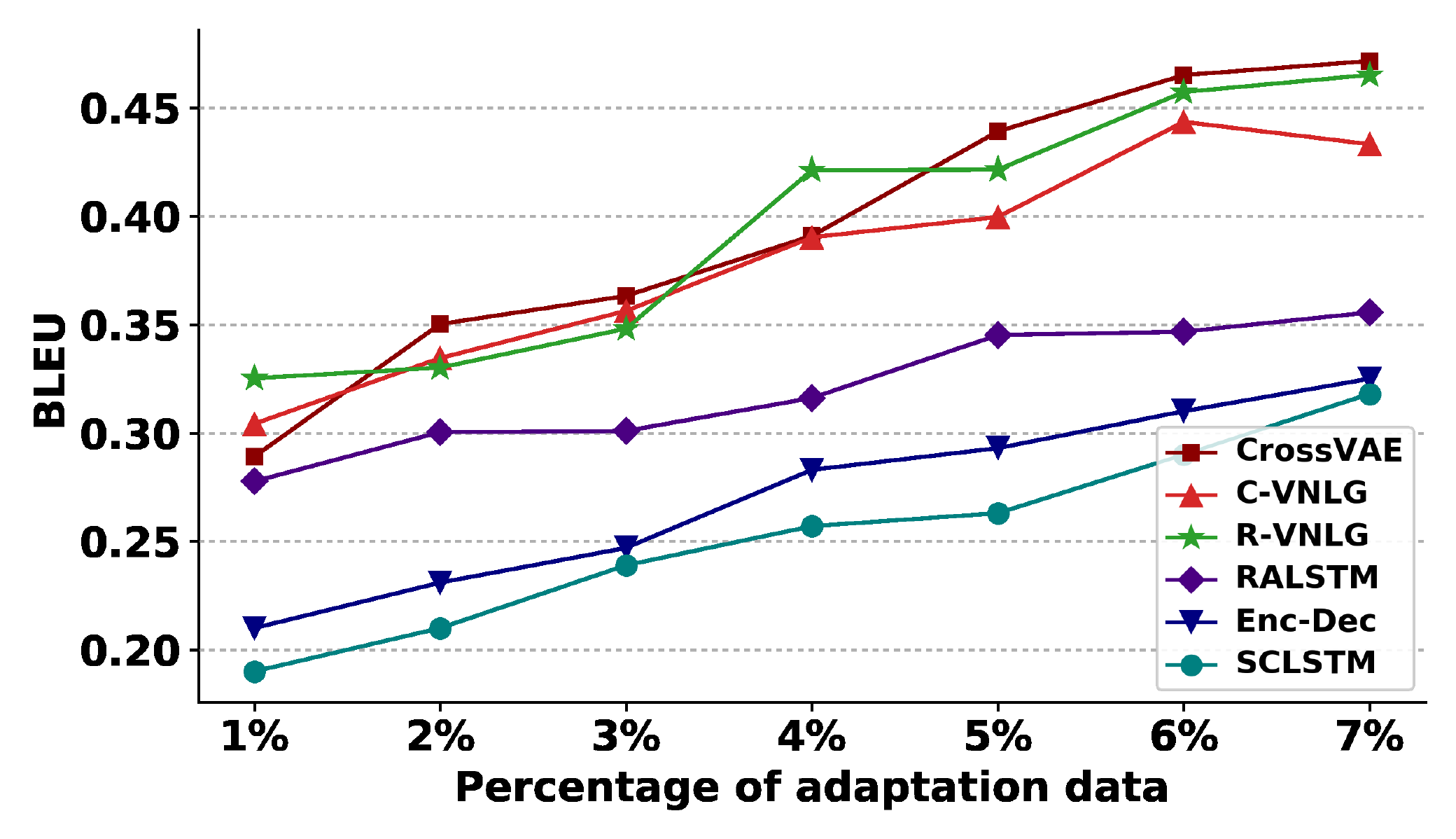}
     
    \caption{Performance on Laptop domain with varied limited amount, from 1\% to 7\%, of the adaptation training data when adapting models pre-trained on [Restaurant+Hotel] union dataset.}
    \label{fig:cVariational-line_chart-integrating}
     
\end{figure}

\section{Results and Analysis}\label{sec:cVariational-resultsanalysis}
We performed the models in different scenarios as follows: (i) \textit{scratch} training where models trained from scratch using $10$\% (\textit{scr10}), $30$\% (\textit{scr30}), and $100$\% (\textit{scr100}) amount of in-domain data; and (ii) domain \textit{adaptation} training where models pre-trained from scratch using all source domain data, then fine-tuned on the target domain using only $10\%$ amount of the target data. Overall, the proposed models can work well in scenarios of low-resource setting data. The proposed models obtained state-of-the-art performances regarding both the evaluation metrics across all domains in all training scenarios. 
\begin{table*}[!ht]
\centering
\resizebox{0.85\textwidth}{!}{%
\begin{tabular}{c|ccccccccc}
\hline 
&\multirow{2}{*}{Model} & \multicolumn{2}{c}{\textbf{Hotel}} & \multicolumn{2}{c}{\textbf{Restaurant}} & \multicolumn{2}{c}{\textbf{Tv}} 		& \multicolumn{2}{c}{\textbf{Laptop}} \\ \cline{3-10} 
&							& BLEU 	  & ERR 	& BLEU   & ERR     & BLEU    & ERR     & BLEU   & ERR     \\ 
\hline
\parbox[t]{2mm}{\multirow{8}{*}{\rotatebox[origin=c]{90}{\textbf{scr100}}}}
&HLSTM  	& 0.8488  & 2.79\%  & 0.7436 & 0.85\%  & 0.5240  & 2.65\%  & 0.5130 & 1.15\% \\
&SCLSTM  	& 0.8469  & 3.12\%  & 0.7543 & 0.57\%  & 0.5235  & 2.41\%  & 0.5109 & 0.89\% \\
&ENCDEC     & 0.8537  & 4.78\%  & 0.7358 & 2.98\%  & 0.5142  & 3.38\%  & 0.5101 & 4.24\% \\
&RALSTM 	& \textbf{0.8965}  & 0.58\%  & \textbf{\textit{0.7779}} & \textbf{\textit{0.20}}\%  & \textbf{\textit{0.5373}}  	& \textbf{\textit{0.49}}\%  & \textbf{\textit{0.5231}} & \textbf{0.50}\% \\\cline{3-10}
&R-VNLG (Ours)   		& 0.8851  & 0.57\%  & 0.7709 & 0.36\%  & 0.5356  & 0.73\%  & 0.5210 & 0.59\% \\
&C-VNLG (Ours)   		& 0.8811  & \textbf{\textit{0.49}}\%  & 0.7651 & \textbf{0.06}\%  & 0.5350  & 0.88\%  & 0.5192 & \textbf{\textit{0.56}}\% \\
&DualVAE (Ours)   		& 0.8813  & \textbf{0.33}\%  & 0.7695 & 0.29\%  & 0.5359  & 0.81\%  & 0.5211 & 0.91\% \\
&CrossVAE (Ours)   		& \textbf{\textit{0.8926}}  & 0.72\%  & \textbf{0.7786} & 0.54\%  & \textbf{0.5383}  & \textbf{0.48}\%  & \textbf{0.5240} & \textbf{0.50}\% \\ 
\hline
\parbox[t]{2mm}{\multirow{8}{*}{\rotatebox[origin=c]{90}{\textbf{scr10}}}}
&HLSTM  	& 0.7483  & 8.69\%  & 0.6586 & 6.93\%  & 0.4819  & 9.39\%  & 0.4813 & 7.37\% \\
&SCLSTM  	& 0.7626  & 17.42\%  & 0.6446 & 16.93\%  & 0.4290  & 31.87\%  & 0.4729 & 15.89\% \\
&ENCDEC     & 0.7370  & 23.19\%  & 0.6174 & 23.63\%  & 0.4570  & 21.28\%  & 0.4604 & 29.86\% \\
&RALSTM 	& 0.6855  & 22.53\%  & 0.6003 & 17.65\%  & 0.4009  & 22.37\% & 0.4475 & 24.47\% \\ \cline{3-10}
&R-VNLG (Ours)		& 0.7378  & 15.43\% & 0.6417 & 15.69\% & 0.4392  & 17.45\% & 0.4851	& 10.06\% \\
&C-VNLG (Ours)   	& 0.7998  & 8.67\%  & 0.6838 & \textbf{\textit{6.86}}\%  & 0.5040  & 5.31\%  & 0.4932 & 3.56\% \\
&DualVAE (Ours)   		& \textbf{\textit{0.8022}}  & \textbf{\textit{6.61}}\%  & \textbf{\textit{0.6926}} & 7.69\%  & \textbf{\textit{0.5110}}  & \textbf{\textit{3.90}}\%  & \textbf{\textit{0.5016}} & \textbf{\textit{2.44}}\% \\
&CrossVAE (Ours)   			& \textbf{0.8103}  & \textbf{6.20}\%  & \textbf{0.6969} & \textbf{4.06}\%  & \textbf{0.5152}  & \textbf{2.86}\%  & \textbf{0.5085} & \textbf{2.39}\% \\ 
\hline
\parbox[t]{2mm}{\multirow{7}{*}{\rotatebox[origin=c]{90}{\textbf{scr30}}}}
& HLSTM 			& 0.8104  & 6.39\%  & 0.7044 & 2.13\%  & 0.5024  & 5.82\%  & 0.4859 & 6.70\% \\
& SCLSTM 			& 0.8271  & 6.23\%  & 0.6825 & 4.80\%  & 0.4934  & 7.97\%  & 0.5001 & 3.52\% \\
& ENCDEC    		& 0.7865  & 9.38\%  & 0.7102 & 13.47\% & 0.5014  & 9.19\%  & 0.4907 & 10.72\% \\
& RALSTM			& 0.8334  & 4.23\%  & 0.7145 & 2.67\%  & 0.5124  & 3.53\%  & 0.5106 & 2.22\% \\ \cline{3-10}
& C-VNLG (Ours) & \textbf{\textit{0.8553}}  & 2.64\%  & 0.7256 & \textbf{\textit{0.96}}\%  & 0.5265  & \textbf{0.66}\%  & \textbf{\textit{0.5117}} & 2.15\% \\
& DualVAE (Ours)   	& 0.8534  & \textbf{\textit{1.54}}\%  & \textbf{\textit{0.7301}} & 2.32\%  & \textbf{\textit{0.5288}}  & 1.05\%  & 0.5107 & \textbf{\textit{0.93}}\% \\
& CrossVAE (Ours)   	& \textbf{0.8585}  & \textbf{1.37}\%  & \textbf{0.7479} & \textbf{0.49}\%  & \textbf{0.5307}  & \textbf{\textit{0.82}}\%  & \textbf{0.5154} & \textbf{0.81}\% \\ 
\hline
\end{tabular}%
}

\caption[Results comparison on a variety of scratch training]{Results evaluated on four domains by training models from \textit{scratch} with \textit{10\%}, \textit{30\%}, and \textit{100\%} in-domain data, respectively. The results were averaged over 5 randomly initialized networks. The \textbf{bold} and \textbf{\textit{italic}} faces denote the best and second best models in each training scenario, respectively.}  
\label{tab:cVariational-DualVAE-results}
\end{table*}

\subsection{Integrating Variational Inference}\label{subsec:cVariational-variationalintegrating} 
We compare the encoder-decoder RALSTM model to its modification by integrating with variational inference (R-VNLG and C-VNLG) as demonstrated in Figure \ref{fig:cVariational-line_chart-integrating} and Table \ref{tab:cVariational-DualVAE-results}. 

It clearly shows that the variational generators not only provide a compelling evidence on adapting to a new, unseen domain when the target domain data is scarce, \textit{i.e}., from $1$\% to $7$\% (Figure \ref{fig:cVariational-line_chart-integrating}) but also preserve the power of the original RALSTM on generation task since their performances are very competitive to those of RALSTM (Table \ref{tab:cVariational-DualVAE-results}, \textit{scr100}). 
Table \ref{tab:cVariational-DualVAE-results}, \textit{scr10} further shows the necessity of the integrating in which the VNLGs achieved a significant improvement over the RALSTM in \textit{scr10} scenario where the models trained from \textit{scratch} with only a limited amount of training data ($10$\%).
These indicate that the proposed variational method can learn the underlying semantic of the existing DA-utterance pairs, which are especially useful information for low-resource setting.

Furthermore, the R-VNLG model has slightly better results than the C-VNLG when providing sufficient training data in \textit{scr100}. 
In contrast, with a modest training data, in \textit{scr10}, the latter model demonstrates a significant improvement compared to the former in terms of both the BLEU and ERR scores by a large margin across all four dataset. 
Take Hotel domain, for example, the C-VNLG model ($79.98$ BLEU, $8.67\%$ ERR) has better results in comparison to the R-VNLG ($73.78$ BLEU, $15.43\%$ ERR) and RALSTM ($68.55$ BLEU, $22.53\%$ ERR).
Thus, the rest experiments focus on the C-VNLG since it shows obvious sign for constructing a dual latent variable models dealing with low-resource in-domain data. We leave the R-VNLG for future investigation.

\begin{figure*}[!ht]  
	\centering
    \vspace{-15pt}
    \includegraphics[width=0.9\textwidth, height=.22\textwidth]{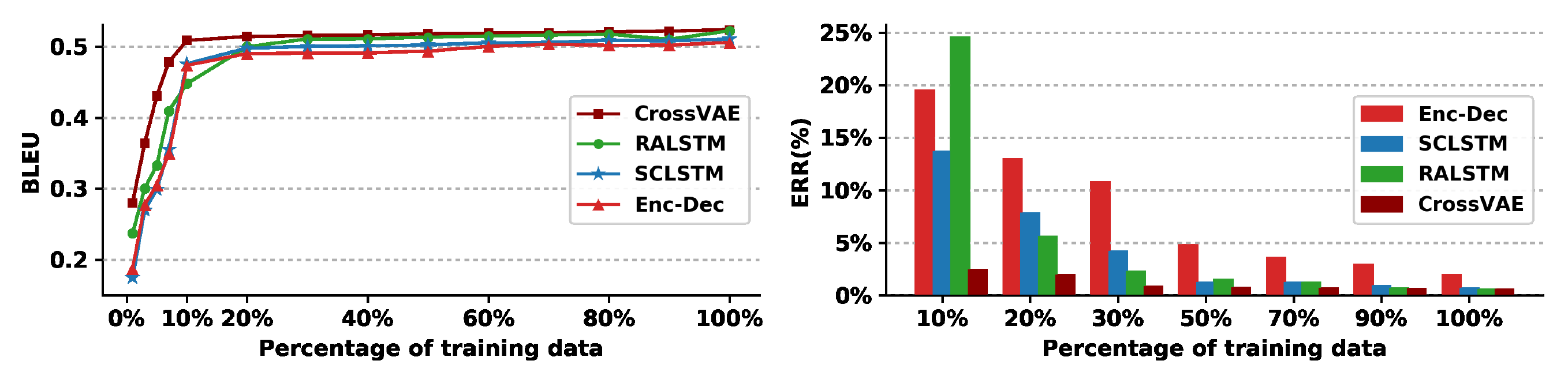}
    \vspace{-10pt}
    \caption{Performance comparison of the models trained on Laptop domain.}
    \label{fig:cVariational-DualVAE-laptop-10to100}
    
\end{figure*}

\subsection{Ablation Studies}\label{subsubsec:cVariational-DualVAE-ablations}
The ablation studies (Table \ref{tab:cVariational-DualVAE-results}) demonstrate the contribution of each model components, in which we incrementally train the baseline RALSTM, the C-VNLG (= RALSTM + Variational inference), the DualVAE (= C-VNLG + Variational CNN-DCNN), and the CrossVAE (= DualVAE + Cross training) models. Generally, while all models can work well when there are sufficient training datasets, the performances of the proposed models also increase as increasing the model components. The trend is consistent across all training cases no matter how much the training data was provided. Take, for example, the \textit{scr100} scenario in which the CrossVAE model mostly outperformed all the previous strong baselines with regard to the BLEU and the slot error rate ERR scores. 

On the other hand, the previous methods showed extremely impaired performances regarding low BLEU score and high slot error rate ERR when training the models from \textit{scratch} with only $10\%$ of in-domain data (\textit{scr10}). 
In contrast, by integrating the variational inference, the C-VNLG model, for example in Hotel domain, can significantly improve the BLEU score from $68.55$ to $79.98$, and also reduce the slot error rate ERR by a large margin, from $22.53$ to $8.67$, compared to the RALSTM baseline. Moreover, the proposed models have much better performance over the previous ones in the \textit{scr10} scenario since the CrossVAE, and the DualVAE models mostly obtained the best and second best results, respectively. The CrossVAE model trained on \textit{scr10} scenario, in some cases, achieved results which close to those of the HLSTM, SCLSTM, and ENCDEC models trained on all training data (\textit{scr100}) scenario. 
Take, for example, the most challenge dataset Laptop, in which the DualVAE and CrossVAE obtained competitive results regarding the BLEU score, at $50.16$ and $50.85$ respectively, which close to those of the HLSTM ($51.30$ BLEU), SCLSTM ($51.09$ BLEU), and ENCDEC ($51.01$ BLEU), while the results regardless the slot error rate ERR scores are also close to those of the previous or even better in some cases, for example DualVAE ($2.44$ ERR), CrossVAE ($2.39$ ERR), and ENCDEC ($4.24$ ERR). There are also some cases in TV domain where the proposed models (in \textit{scr10}) have results close to or better over the previous ones (trained on \textit{scr100}).   
These indicate that the proposed models can encode useful information into the latent variable efficiently to better generalize to the unseen dialogue acts, addressing the second difficulty with low-resource data.

The \textit{scr30} section further confirms the effectiveness of the proposed methods, in which the CrossVAE and DualVAE still mostly rank the best and second-best models compared with the baselines. The proposed models also show superior ability in leveraging the existing small training data to obtain very good performances, which are in many cases even better than those of the previous methods trained on $100$\% of in-domain data. Take Tv domain, for example, in which the CrossVAE in \textit{scr30} achieves a good result regarding BLEU and slot error rate ERR score, at $53.07$ BLEU and $0.82$ ERR, that are not only competitive to the RALSTM ($53.73$ BLEU, $0.49$ ERR), but also outperform the previous models in \textit{scr100} training scenario, such as HLSTM ($52.40$ BLEU, $2.65$ ERR), SCLSTM ($52.35$ BLEU, $2.41$ ERR), and ENCDEC ($51.42$ BLEU, $3.38$ ERR). This further indicates the need of the integrating with variational inference, the additional auxiliary auto-encoding, as well as the joint and cross training.

\subsection{Model comparison on unseen domain}\label{subsubsec:cVariational-DualVAE-unseendomain}
In this experiment, we trained four models (ENCDEC, SCLSTM, RALSTM, and CrossVAE) from \textit{scratch} in the most difficult unseen Laptop domain with an increasingly varied proportion of training data, start from $1$\% to $100$\%. The results are shown in Figure \ref{fig:cVariational-DualVAE-laptop-10to100}. It clearly sees that the BLEU score increases and the slot error ERR decreases as the models are trained on more data. The CrossVAE model is clearly better than the previous models (ENCDEC, SCLSTM, RALSTM) in all cases. While the performance of the CrossVAE, RALSTM model starts to saturate around $30$\% and $50$\%, respectively, the ENCDEC model seems to continue getting better as providing more training data. The figure also confirms that the CrossVAE trained on $30$\% of data can achieve a better performance compared to those of the previous models trained on $100$\% of in-domain data.

\begin{table*}[!ht]  
\centering

\resizebox{0.8\textwidth}{!}{%
\begin{tabular}{ccccccccc}
\hline 
\multirow{2}{*}{\diagbox{Source}{\textbf{Target}}} & \multicolumn{2}{c}{\textbf{Hotel}} & \multicolumn{2}{c}{\textbf{Restaurant}} & \multicolumn{2}{c}{\textbf{Tv}} 		& \multicolumn{2}{c}{\textbf{Laptop}} \\ \cline{2-9} 
						& BLEU 	  & ERR 	& BLEU   & ERR     & BLEU    & ERR     & BLEU   & ERR   \\ 
\hline
Hotel$^\flat$  			& - 	  & - 	    & 0.6243 & 11.20\% & 0.4325  & 29.12\% & 0.4603 & 22.52\% \\
Restaurant$^\flat$   	& 0.7329  & 29.97\%  & - 	 & - 	   & 0.4520  & 24.34\% & 0.4619 & 21.40\% \\
Tv$^\flat$ 				& 0.7030  & 25.63\% & 0.6117 & 12.78\% & - 	 	 & - 	   & 0.4794 & 11.80\% \\
Laptop$^\flat$ 			& 0.6764  & 39.21\% & 0.5940 & 28.93\% & 0.4750  & 14.17\% & - 		& - 	\\  
\hline 
Hotel$^{\sharp}$  		& - 	  & - 	    & 0.7138 & 2.91\%  & 0.5012  & 5.83\%  & 0.4949 & 1.97\% \\
Restaurant$^{\sharp}$   & 0.7984  & 4.04\%  & - 	 & - 	   & 0.5120  & 3.26\%  & 0.4947 & 1.87\% \\
Tv$^{\sharp}$ 			& 0.7614  & 5.82\%  & 0.6900 & 5.93\%  & - 	 	 & - 	   & 0.4937 & 1.91\% \\
Laptop$^{\sharp}$ 		& 0.7804  & 5.87\%  & 0.6565 & 6.97\%  & 0.5037  & 3.66\%  & -      & - 	\\ 
\hline 
Hotel$^{\xi}$  			& - 	  & - 	    & 0.6926 & 3.56\%  & 0.4866  & 11.99\%  & 0.5017 & 3.56\% \\
Restaurant$^{\xi}$   	& 0.7802  & 3.20\%  & - 	 & - 	   & 0.4953  & 3.10\%   & 0.4902 & 4.05\% \\
Tv$^{\xi}$ 				& 0.7603  & 8.69\%  & 0.6830 & 5.73\%  & - 	 	 & - 	    & 0.5055 & 2.86\% \\
Laptop$^{\xi}$ 			& 0.7807  & 8.20\%  & 0.6749 & 5.84\%  & 0.4988  & 5.53\%  & -      & - 	\\ 
\hline 
CrossVAE (\textit{scr10})   			& 0.8103  & 6.20\%  & 0.6969 & 4.06\%  & 0.5152  & 2.86\%  & 0.5085 & 2.39\% 
\\ \hline
CrossVAE (\textit{semi-U50-L10})   			& 0.8144  & 6.12\%  & 0.6946 & 3.94\%  & 0.5158  & 2.95\%  & 0.5086 & 1.31\% 
\\ \hline
\end{tabular}%
}
\vspace{-5pt}
\caption{Results evaluated on \textbf{Target} domains by \textit{adaptation} training SCLSTM model from $100$\% (denoted as \textbf{$\flat$}) of Source data, and the CrossVAE model from 30\% (denoted as \textbf{$\sharp$}), 100\% (denoted as \textbf{$\xi$}) of Source data. The scenario used only 10\% amount of the \textbf{Target} domain data. 
The last two rows show results by training the CrossVAE model on the \textit{scr10} and semi-supervised learning, respectively. 
} 
\label{tab:cVariational-DualVAE-adaptation-semi}
\vspace{-5pt}
\end{table*}

\subsection{Domain Adaptation}\label{subsubsec:cVariational-DualVAE-domainadaptation}
We further examine the domain scalability of the proposed methods by training the CrossVAE and SCLSTM models on \textit{adaptation} scenarios, in which we first trained the models on out-of-domain data, and then fine-tuned the model parameters by using a small amount ($10$\%) of in-domain data. The results are shown in Table \ref{tab:cVariational-DualVAE-adaptation-semi}. 

Both SCLSTM and CrossVAE models can take advantage of ``\textit{close}'' dataset pairs, \textit{i.e}., Restaurant $\leftrightarrow$ Hotel, and Tv $\leftrightarrow$ Laptop, to achieve better performances compared to those of the ``\textit{different}'' dataset pairs, \textit{i.e}. Latop $\leftrightarrow$ Restaurant. Moreover, Table \ref{tab:cVariational-DualVAE-adaptation-semi} clearly shows that the SCLSTM (denoted by $\flat$) is limited to scale to another domain in terms of having very low BLEU and high ERR scores. This adaptation scenario along with the \textit{scr10} and \textit{scr30} in Table \ref{tab:cVariational-DualVAE-results} demonstrate that the SCLSTM can not work when having a low-resource setting of in-domain training data.

On the other hand, the CrossVAE model again show ability in leveraging the out-of-domain data to better adapt to a new domain. Especially in the case where Laptop, which is a most difficult unseen domain, is the target domain the CrossVAE model can obtain good results irrespective of low slot error rate ERR, around $1.90$\%, and high BLEU score, around $50.00$ points. Surprisingly, the CrossVAE model trained on \textit{scr10} scenario in some cases achieves better performance compared to those in adaptation scenario first trained with $30$\% out-of-domain data (denoted by $\sharp$) which is also better than the adaptation model trained on $100$\% out-of-domain data (denoted by $\xi$). 

Preliminary experiments on semi-supervised training were also conducted, in which we trained the CrossVAE model with the same $10$\% in-domain \textit{labeled} data as in the other scenarios and $50$\% in-domain \textit{unlabeled} data by keeping only the utterances \textbf{u} in a given input pair of dialogue act-utterance (\textbf{d}, \textbf{u}), denoted by \textit{semi-U50-L10}. The results showed CrossVAE's ability in leveraging the unlabeled data to achieve slightly better results compared to those in \textit{scratch} scenario. 
All these stipulate that the proposed models can perform acceptably well in training cases of \textit{scratch}, domain \textit{adaptation}, and \textit{semi-supervised} where the in-domain training data is in short supply.

\begin{table*}[!ht]
\vspace{-5pt}
\centering
\resizebox{\textwidth}{!}{%
\begin{tabularx}{1.45\textwidth}{p{0.15\textwidth}p{1.25\textwidth}}
\textbf{Model} & \textbf{Generated Responses from TV Domain} \\ \hline 
\textbf{\textit{DA 1}} & compare(name=`\textit{typhon 45}'; hdmiport=`\textit{2}'; family=`\textit{l2}'; name=`\textit{hades 48}'; hdmiport=`\textit{4}'; family=`\textit{l7}') \\
\textbf{\textit{Reference 1}} & Compared to \textit{typhon 45} which has \textit{2 hdmi port -s} and is in the \textit{L2 product family}, \textit{hades 48} has \textit{4 hdmi port -s} and is in the \textit{L7 product family}. Which one do you prefer ? \\ \hline

ENCDEC-10 & the \textit{typhon 45} is in the \textit{l2 product family} and has \textit{2 hdmi port -s} and is in the \textcolor{violet}{\textit{l7 product family}} with \textcolor{violet}{\textit{4 hdmi port -s}}, the \textit{hades 48} is in the \textcolor{orange}{SLOT\_FAMILY} product family with a \textcolor{teal}{\textit{SLOT\_AUDIO}}.
\\ 
HLSTM-10 & the \textit{typhon 45} is a great \textcolor{teal}{\textit{eco rating}}, the \textit{hades 48} is in the \textcolor{violet}{\textit{l2 family}} with \textcolor{violet}{\textit{2 hdmi port -s}}. \textcolor{red}{[l7 family]} \textcolor{red}{[4 hdmi port -s]}
\\
SCLSTM-10 & the \textit{typhon 45} is the \textcolor{violet}{\textit{hades 48}} with \spelling{\textit{2 hdmi port}} in the \textit{l2 family}, the \textcolor{orange}{\textit{SLOT\_NAME}} has \textit{4 hdmi port -s} and \textcolor{orange}{\textit{SLOT\_HDMIPORT}} hdmi port. \textcolor{red}{[l7 family]}
\\
C-VNLG-10 & the \textit{typhon 45} has \textit{2 hdmi port -s} and the \textit{hades 48} is in the \textcolor{violet}{\textit{l2 family}} and has \textit{4 hdmi port -s}. \textcolor{red}{[l7 family]}
\\ 
DualVAE-10 & the \textit{typhon 45} has \textit{2 hdmi port -s} and is in the \textit{l2 family} while the \textit{hades 48} has \textit{4 hdmi port -s} and is in the \textit{l7 family}. \OK{[OK]}
\\
CrossVAE-10 & the \textit{typhon 45} is in the \textit{l2 family} with \textit{2 hdmi port -s} while the \textit{hades 48} has \textit{4 hdmi port -s} and is in the \textit{l7 family}. \OK{[OK]}
\\\hline
ENCDEC-30 & the \textit{typhon 45} has \textit{2 hdmi port -s}, the \textit{hades 48} has \textit{4 hdmi port -s}, the \redundant{\textit{SLOT\_NAME}} has \redundant{\textit{SLOT\_HDMIPORT}} hdmi port. \missing{[l2 family]} \missing{[l7 family]} 
\\ 
HLSTM-30 & the \textit{typhon 45} is in the \textit{l2 product family} with \textit{2 hdmi port -s}, whereas the \textit{hades 48} has \spelling{\textit{4 hdmi port}}. \missing{[l7 family]} 
\\
SCLSTM-30 & the \textit{typhon 45} has \textit{2 hdmi port -s}, the \textit{hades 48} is in the \misplaced{\textit{l2 product family}}. \missing{[l7 family]} \missing{[4 hdmi port -s]}
\\
C-VNLG-30 & the \textit{typhon 45} has \textit{2 hdmi port -s}, the \textit{hades 48} is in the \misplaced{\textit{l2 product family}} and has \textit{4 hdmi port -s} in \textit{l7 family}.
\\ 
DualVAE-30 & which do you prefer, the \textit{typhon 45} in the \textit{l2 product family} with \textit{2 hdmi port -s} . the \textit{hades 48} is in the \textit{l7 family} with \textit{4 hdmi port -s}. \OK{[OK]}
\\
CrossVAE-30 & the \textit{typhon 45} has \textit{2 hdmi port -s} and in the \textit{l2 family} while the \textit{hades 48} has \textit{4 hdmi port -s} and is in the \textit{l7 family}. which item do you prefer. \OK{[OK]}
\\\hline 

\hline
\textbf{\textit{DA 2}} & recommend(name=`\textit{proteus 73}'; type=`\textit{television}'; price=`\textit{1500 dollars}'; audio=`\textit{nicam stereo}'; hdmiport=`\textit{2}') \\
\textbf{\textit{Reference 2}} & \textit{proteus 73} is a nice \textit{television}. its price is \textit{1500 dollars}, its audio is \textit{nicam stereo}, and it has \textit{2 hdmi port -s}. \\ \hline

ENCDEC-10 & the \textit{proteus 73} is a great \textit{television} with a \textit{nicam stereo} and \textit{2 hdmi port -s} \missing{[1500 dollars]}
\\ 
HLSTM-10 & the \textit{proteus 73} is a \textit{television} with \textit{2 hdmi port -s} and comes with a \textit{nicam stereo} and \textit{costs 1500 dollars} \OK{[OK]}
\\
SCLSTM-10 & the \textit{proteus 73} is a nice \textit{television} with \textit{nicam stereo} and \textit{2 hdmi port -s} \missing{[1500 dollars]}
\\
C-VNLG-10 & the \textit{proteus 73} \textit{television} has a \textit{nicam stereo} and \textit{2 hdmi port -s} and \textit{costs 1500 dollars} \OK{[OK]}
\\ 
DualVAE-10 & the \textit{proteus 73} \textit{television} has a \textit{nicam stereo} and \textit{2 hdmi port -s} and \textit{costs 1500 dollars} \OK{[OK]}
\\
CrossVAE-10 & the \textit{proteus 73} \textit{television} has \textit{2 hdmi port -s} and a \textit{nicam stereo} and \textit{costs 1500 dollars} \OK{[OK]}
\\\hline
ENCDEC-30 & the \textit{proteus 73} \textit{television} has \textit{2 hdmi port -s} and \textit{nicam stereo audio} for \textit{1500 dollars} \OK{[OK]}
\\ 
HLSTM-30 & the \textit{proteus 73} \textit{television} has a \textit{nicam stereo} and \textit{2 hdmi port -s} and is priced at \textit{1500 dollars} \OK{[OK]}
\\
SCLSTM-30 & the \textit{proteus 73} is a nice \textit{television} with \textit{nicam stereo} and \textit{2 hdmi port -s} . it is priced at \textit{1500 dollars} \OK{[OK]}
\\
C-VNLG-30 & the \textit{proteus 73} \textit{television} has \textit{2 hdmi port -s} , \textit{nicam stereo audio} , and \textit{costs 1500 dollars} \OK{[OK]}
\\ 
DualVAE-30 & the \textit{proteus 73} \textit{television} has \textit{2 hdmi port -s} and \textit{nicam stereo audio} and \textit{costs 1500 dollars} \OK{[OK]}
\\
CrossVAE-30 & the \textit{proteus 73} \textit{television} has \textit{2 hdmi port -s} and \textit{nicam stereo audio} and \textit{costs 1500 dollars} \OK{[OK]}
\\\hline

\end{tabularx} %
}  
\vspace{-5pt}
\caption{Comparison of top \textbf{Tv} responses generated for different models in different scenarios. Errors are marked in colors (\textcolor{red}{[missing]}, \textcolor{violet}{misplaced}, \textcolor{orange}{redundant}, \textcolor{teal}{wrong}, \textcolor{olive}{spelling mistake} information). \textcolor{blue}{[OK]} denotes successful generation. Model-X where X is amount of training data, \textit{i.e}. 10\%, 30\%, or 100\%.} 
\label{tab:cVariational-DualVAE-TV-output-1}
\end{table*}

\begin{table*}[!ht]
\centering
\resizebox{\textwidth}{!}{%
\begin{tabularx}{1.45\textwidth}{p{0.15\textwidth}p{1.25\textwidth}}
\textbf{Model} & \textbf{Generated Responses from Laptop Domain} \\ \hline 
\textbf{\textit{DA}} & compare(name=`\textit{satellite pallas 21}'; battery=`\textit{4 hour}'; drive=`\textit{500 gb}'; name=`\textit{satellite dinlas 18}'; battery=`\textit{3.5 hour}'; drive=`\textit{1 tb}') \\
\textbf{\textit{Reference}} & compared to satellite \textit{pallas 21} which can last \textit{4 hour} and has a \textit{500 gb drive} , \textit{satellite dinlas 18} can last \textit{3.5 hour} and has a \textit{1 tb drive} . which one do you prefer
\\ \hline
Enc-Dec-10 & the \textit{satellite pallas 21} has a \textit{500 gb drive} , the \textit{satellite dinlas 18} has a \misplaced{\textit{4 hour battery}} life and a \textit{3.5 hour battery} life and a \redundant{\textit{SLOT\_BATTERY}} battery life and a \textit{1 tb drive}
\\ 
HLSTM-10 & the \textit{satellite pallas 21} has a \textit{4 hour battery} life and a \textit{500 gb drive} . which one do you prefer \missing{[satellite pallas 18]} \missing{[3.5 hour battery]} \missing{[1 tb drive]}
\\
SCLSTM-10 & the \textit{satellite pallas 21} has a \textit{4 hour battery} , and has a \misplaced{\textit{3.5 hour battery}} life and a \textit{500 gb drive} and a \misplaced{\textit{1 tb drive}} \missing{[satellite dinlas 18]}
\\
C-VNLG-10 & the \textit{satellite pallas 21} has a \textit{500 gb drive} and a \textit{4 hour battery} life . the \textit{satellite dinlas 18} has a \textit{3.5 hour battery} life and a \redundant{\textit{SLOT\_BATTERY}} battery life \missing{[1 tb drive]}
\\ 
DualVAE-10 & the \textit{satellite pallas 21} has a \textit{4 hour battery} life and a \textit{500 gb drive} and the \textit{satellite dinlas 18} with a \textit{3.5 hour battery} life and \wrong{\textit{is good for business computing}} . which one do you prefer \missing{[1 tb drive]}
\\
CrossVAE-10 & the \textit{satellite pallas 21} with \textit{500 gb} and a \misplaced{\textit{1 tb drive}} . the \textit{satellite dinlas 18} with a \misplaced{\textit{4 hour battery}} and a \redundant{\textit{SLOT\_DRIVE}} drive . which one do you prefer \missing{[3.5 hour battery]}
\\\hline
Enc-Dec-30 & the \textit{satellite pallas 21} has a \textit{500 gb drive} with a \misplaced{\textit{1 tb drive}} and is the \misplaced{\textit{satellite dinlas 18}} with a \redundant{\textit{SLOT\_DRIVE}} drive for \misplaced{\textit{4 hour -s}} . which one do you prefer \missing{[3.5 hour battery]}
\\ 
HLSTM-30 & the \textit{satellite pallas 21} is a \textit{500 gb drive} with a \textit{4 hour battery} life . the \textit{satellite dinlas 18} has a \textit{3.5 hour battery} life . which one do you prefer \missing{[1 tb drive]}
\\
SCLSTM-30 & the \textit{satellite pallas 21} has a \textit{500 gb drive} . the \textit{satellite dinlas 18} has a \misplaced{\textit{4 hour battery}} life . the \redundant{\textit{SLOT\_NAME}} has a \textit{3.5 hour battery} life . which one do you prefer \missing{[1 tb drive]}
\\
C-VNLG-30 & which one do you prefer the \textit{satellite pallas 21} with a \textit{4 hour battery} life , the \textit{satellite dinlas 18} has a \misplaced{\textit{500 gb drive}} and a \textit{3.5 hour battery} life and a 1 tb drive . which one do you prefer 
\\ 
DualVAE-30 & \textit{satellite pallas 21} has a \textit{500 gb drive} and a \textit{4 hour battery} life while the \textit{satellite dinlas 18} with a \textit{3.5 hour battery} life and a \textit{1 tb drive} . \OK{[OK]}
\\
CrossVAE-30 & the \textit{satellite pallas 21} has a \textit{500 gb drive} with a \textit{4 hour battery} life . the \textit{satellite dinlas 18} has a \textit{1 tb drive} and a \textit{3.5 hour battery} life . which one do you prefer \OK{[OK]}
\\\hline
\end{tabularx} %
}
\vspace{-5pt}
\caption{Comparison of top \textbf{Laptop} responses generated for different models in different scenarios. Errors are marked in colors (\textcolor{red}{[missing]}, \textcolor{violet}{misplaced}, \textcolor{orange}{redundant}, \textcolor{teal}{wrong}, \textcolor{olive}{spelling} information). \textcolor{blue}{[OK]} denotes successful generation. Model-X where X is amount of training data, \textit{i.e}. 10\%, 30\%, or 100\%.}
\label{tab:lap-comparison}
\end{table*}

\subsection{Comparison on Generated Outputs}\label{subsubsec:cVariational-DualVAE-outputs}
We present top responses generated for different scenarios from TV (Table \ref{tab:cVariational-DualVAE-TV-output-1}) and Laptop (Table \ref{tab:lap-comparison}), which further show the effectiveness of the proposed methods. 

On the one hand, previous models trained on \textit{scr10}, \textit{scr30} scenarios produce a diverse range of the outputs' error types, including \missing{missing}, \misplaced{misplaced}, \redundant{redundant}, \wrong{wrong} slots, or \spelling{spelling mistake} information, resulting in a very high score of the slot error rate ERR. The ENCDEC, HLSTM and SCLSTM models in Table \ref{tab:cVariational-DualVAE-TV-output-1}-\textbf{DA 1}, for example, tend to generate outputs with redundant slots (\textit{i.e.}, \redundant{\textit{SLOT\_HDMIPORT}}, \textcolor{orange}{\textit{SLOT\_NAME}}, \textcolor{orange}{SLOT\_FAMILY}), missing slots (\textit{i.e.}, \textcolor{red}{[l7 family]}, \textcolor{red}{[4 hdmi port -s]}), or even in some cases produce irrelevant slots (\textit{i.e.}, \textcolor{teal}{\textit{SLOT\_AUDIO}}, \textcolor{teal}{\textit{eco rating}}), resulting in inadequate utterances. 

On the other hand, the proposed models can effectively leverage the knowledge from only few of the existing training instances to better generalize to the unseen dialogue acts, leading to satisfactory responses. For example in Table \ref{tab:cVariational-DualVAE-TV-output-1}, the proposed methods can generate adequate number of the required slots, resulting in fulfilled utterances (DualVAE-10, CrossVAE-10, DualVAE-30, CrossVAE-30), or acceptable outputs with much fewer error information, \textit{i.e.}, mis-ordered slots in the generated utterances (C-VNLG-30).

For a much easier dialogue act in Table \ref{tab:cVariational-DualVAE-TV-output-1}-\textbf{DA 2}, previous models still produce some error outputs, whereas the proposed methods seem to form some specific slots into phrase in concise outputs. For example, instead of generating ``the \textit{\underline{proteus 73}} is a \underline{\textit{television}}'' phrase, the proposed models tend to concisely produce ``the \textit{\underline{proteus 73}} \textit{\underline{television}}''. The trend is mostly consistent to those in Table \ref{tab:lap-comparison}.

\section{Conclusion and Future Work}
\label{sec:conclusion_future_work} \vspace{-6pt}
We present an approach to low-resource NLG by integrating the variational inference and introducing a novel auxiliary auto-encoding. 
Experiments showed that the models can perform acceptably well using a scarce dataset. The ablation studies demonstrate that the variational generator contributes to learning the underlying semantic of DA-utterance pairs, while the variational CNN-DCNN plays an important role of encoding useful information into the latent variable. 
In the future, we further investigate the proposed models with adversarial training, semi-supervised, or unsupervised training.
\vspace{-5pt}
\section*{Acknowledgements}
This work was supported by the JST CREST Grant Number JPMJCR1513, the JSPS KAKENHI Grant number 15K16048  and the grant of a collaboration between JAIST and TIS. 

\bibliography{conll2018}

\begin{thebibliography}{19}
\expandafter\ifx\csname natexlab\endcsname\relax\def\natexlab#1{#1}\fi

\bibitem[{Bengio et~al.(2015)Bengio, Vinyals, Jaitly, and
  Shazeer}]{bengio2015scheduled}
Samy Bengio, Oriol Vinyals, Navdeep Jaitly, and Noam Shazeer. 2015.
\newblock Scheduled sampling for sequence prediction with recurrent neural
  networks.
\newblock In \emph{Advances in Neural Information Processing Systems}, pages
  1171--1179.

\bibitem[{Bowman et~al.(2015)Bowman, Vilnis, Vinyals, Dai, J{\'{o}}zefowicz,
  and Bengio}]{DBLP:journals/corr/BowmanVVDJB15}
Samuel~R. Bowman, Luke Vilnis, Oriol Vinyals, Andrew~M. Dai, Rafal
  J{\'{o}}zefowicz, and Samy Bengio. 2015.
\newblock Generating sentences from a continuous space.
\newblock \emph{CoRR}, abs/1511.06349.

\bibitem[{Chung et~al.(2016)Chung, Ahn, and Bengio}]{chung2016hierarchical}
Junyoung Chung, Sungjin Ahn, and Yoshua Bengio. 2016.
\newblock Hierarchical multiscale recurrent neural networks.
\newblock \emph{arXiv preprint arXiv:1609.01704}.

\bibitem[{Chung et~al.(2015)Chung, Kastner, Dinh, Goel, Courville, and
  Bengio}]{chung2015recurrent}
Junyoung Chung, Kyle Kastner, Laurent Dinh, Kratarth Goel, Aaron~C Courville,
  and Yoshua Bengio. 2015.
\newblock A recurrent latent variable model for sequential data.
\newblock In \emph{Advances in neural information processing systems}, pages
  2980--2988.

\bibitem[{Kingma and Welling(2013)}]{kingma2013auto}
Diederik~P Kingma and Max Welling. 2013.
\newblock Auto-encoding variational bayes.
\newblock \emph{arXiv preprint arXiv:1312.6114}.

\bibitem[{Levin et~al.(2000)Levin, Narayanan, Pieraccini, Biatov, Bocchieri,
  Fabbrizio, Eckert, Lee, Pokrovsky, Rahim et~al.}]{levin2000t}
Esther Levin, Shrikanth Narayanan, Roberto Pieraccini, Konstantin Biatov,
  Enrico Bocchieri, Giuseppe~Di Fabbrizio, Wieland Eckert, Sungbok Lee,
  A~Pokrovsky, Mazin Rahim, et~al. 2000.
\newblock The at\&t-darpa communicator mixed-initiative spoken dialog system.
\newblock In \emph{Sixth International Conference on Spoken Language
  Processing}.

\bibitem[{Mairesse et~al.(2010)Mairesse, Ga\v{s}i\'{c}, Jur\v{c}\'{\i}\v{c}ek,
  Keizer, Thomson, Yu, and Young}]{Mairesse:2010:PSL:1858681.1858838}
Fran\c{c}ois Mairesse, Milica Ga\v{s}i\'{c}, Filip Jur\v{c}\'{\i}\v{c}ek, Simon
  Keizer, Blaise Thomson, Kai Yu, and Steve Young. 2010.
\newblock Phrase-based statistical language generation using graphical models
  and active learning.
\newblock In \emph{Proceedings of the 48th Annual Meeting of the Association
  for Computational Linguistics}, ACL '10, pages 1552--1561, Stroudsburg, PA,
  USA. Association for Computational Linguistics.

\bibitem[{Semeniuta et~al.(2017)Semeniuta, Severyn, and
  Barth}]{semeniuta2017hybrid}
Stanislau Semeniuta, Aliaksei Severyn, and Erhardt Barth. 2017.
\newblock A hybrid convolutional variational autoencoder for text generation.
\newblock \emph{arXiv preprint arXiv:1702.02390}.

\bibitem[{Shen et~al.(2017)Shen, Zhang, Henao, Su, and
  Carin}]{shen2017deconvolutional}
Dinghan Shen, Yizhe Zhang, Ricardo Henao, Qinliang Su, and Lawrence Carin.
  2017.
\newblock Deconvolutional latent-variable model for text sequence matching.
\newblock \emph{arXiv preprint arXiv:1709.07109}.

\bibitem[{Sohn et~al.(2015)Sohn, Lee, and Yan}]{sohn2015learning}
Kihyuk Sohn, Honglak Lee, and Xinchen Yan. 2015.
\newblock Learning structured output representation using deep conditional
  generative models.
\newblock In \emph{Advances in Neural Information Processing Systems}, pages
  3483--3491.

\bibitem[{Tran and Nguyen(2017)}]{tran-nguyen:2017:CoNLL}
Van-Khanh Tran and Le-Minh Nguyen. 2017.
\newblock Natural language generation for spoken dialogue system using rnn
  encoder-decoder networks.
\newblock In \emph{Proceedings of the 21st Conference on Computational Natural
  Language Learning (CoNLL 2017)}, pages 442--451, Vancouver, Canada.
  Association for Computational Linguistics.

\bibitem[{Tran et~al.(2017)Tran, Nguyen, and
  Tojo}]{tran-nguyen-tojo:2017:W17-55}
Van-Khanh Tran, Le-Minh Nguyen, and Satoshi Tojo. 2017.
\newblock Neural-based natural language generation in dialogue using rnn
  encoder-decoder with semantic aggregation.
\newblock In \emph{Proceedings of the 18th Annual SIGdial Meeting on Discourse
  and Dialogue}, pages 231--240, Saarbrücken, Germany. Association for
  Computational Linguistics.

\bibitem[{Wen et~al.(2015{\natexlab{a}})Wen, Ga{\v{s}}i\'c, Kim,
  Mrk{\v{s}}i\'c, Su, Vandyke, and Young}]{wenthwsjy15}
Tsung-Hsien Wen, Milica Ga{\v{s}}i\'c, Dongho Kim, Nikola Mrk{\v{s}}i\'c,
  Pei-Hao Su, David Vandyke, and Steve Young. 2015{\natexlab{a}}.
\newblock {Stochastic Language Generation in Dialogue using Recurrent Neural
  Networks with Convolutional Sentence Reranking}.
\newblock In \emph{Proceedings SIGDIAL}. Association for Computational
  Linguistics.

\bibitem[{Wen et~al.(2016{\natexlab{a}})Wen, Gasic, Mrksic, Rojas-Barahona, Su,
  Vandyke, and Young}]{wen2016multi}
Tsung-Hsien Wen, Milica Gasic, Nikola Mrksic, Lina~M Rojas-Barahona, Pei-Hao
  Su, David Vandyke, and Steve Young. 2016{\natexlab{a}}.
\newblock Multi-domain neural network language generation for spoken dialogue
  systems.
\newblock \emph{arXiv preprint arXiv:1603.01232}.

\bibitem[{Wen et~al.(2016{\natexlab{b}})Wen, Ga{\v{s}}ic, Mrk{\v{s}}ic,
  Rojas-Barahona, Su, Vandyke, and Young}]{wentoward}
Tsung-Hsien Wen, Milica Ga{\v{s}}ic, Nikola Mrk{\v{s}}ic, Lina~M
  Rojas-Barahona, Pei-Hao Su, David Vandyke, and Steve Young.
  2016{\natexlab{b}}.
\newblock Toward multi-domain language generation using recurrent neural
  networks.

\bibitem[{Wen et~al.(2015{\natexlab{b}})Wen, Ga{\v{s}}i\'c, Mrk{\v{s}}i\'c, Su,
  Vandyke, and Young}]{wensclstm15}
Tsung-Hsien Wen, Milica Ga{\v{s}}i\'c, Nikola Mrk{\v{s}}i\'c, Pei-Hao Su, David
  Vandyke, and Steve Young. 2015{\natexlab{b}}.
\newblock Semantically conditioned lstm-based natural language generation for
  spoken dialogue systems.
\newblock In \emph{Proceedings of EMNLP}. Association for Computational
  Linguistics.

\bibitem[{Yang et~al.(2017)Yang, Hu, Salakhutdinov, and
  Berg-Kirkpatrick}]{yang2017improved}
Zichao Yang, Zhiting Hu, Ruslan Salakhutdinov, and Taylor Berg-Kirkpatrick.
  2017.
\newblock Improved variational autoencoders for text modeling using dilated
  convolutions.
\newblock \emph{arXiv preprint arXiv:1702.08139}.

\bibitem[{{Zhang} et~al.(2016){Zhang}, {Xiong}, {Su}, {Duan}, and
  {Zhang}}]{2016arXiv160507869Z}
B.~{Zhang}, D.~{Xiong}, J.~{Su}, H.~{Duan}, and M.~{Zhang}. 2016.
\newblock {Variational Neural Machine Translation}.
\newblock \emph{ArXiv e-prints}.

\bibitem[{Zhang et~al.(2017)Zhang, Shen, Wang, Gan, Henao, and
  Carin}]{zhang2017deconvolutional}
Yizhe Zhang, Dinghan Shen, Guoyin Wang, Zhe Gan, Ricardo Henao, and Lawrence
  Carin. 2017.
\newblock Deconvolutional paragraph representation learning.
\newblock In \emph{Advances in Neural Information Processing Systems}, pages
  4172--4182.

\end{thebibliography}
\bibliographystyle{conll2018} 
\end{document}